\documentclass[conference]{IEEEtran}
\IEEEoverridecommandlockouts
\usepackage{cite}
\usepackage{amsmath,amssymb,amsfonts}
\usepackage{algorithmic}
\usepackage{graphicx}
\usepackage{textcomp}
\usepackage{xcolor}
\usepackage{url}
\usepackage{multirow}
\usepackage{tabularx}
\def\BibTeX{{\rm B\kern-.05em{\sc i\kern-.025em b}\kern-.08em
    T\kern-.1667em\lower.7ex\hbox{E}\kern-.125emX}}
\begin{document}

\title{Supervised Negative Binomial Classifier for Probabilistic Record Linkage
} 

\author{\IEEEauthorblockN{Harish Kashyap K}
\IEEEauthorblockA{\textit{MCG} \\
Mysore, India \\
harish.k.kashyap@gmail.com}
\and
\IEEEauthorblockN{ Kiran Byadarhaly}
\IEEEauthorblockA{\textit{ MCG} \\
Mysore, India \\
bkiranv@yahoo.com}
\and
\IEEEauthorblockN{ Saumya Shah}
\IEEEauthorblockA{\textit{MCG} \\
Mysore, India \\
saumya.shah@voyagenius.ai}

}

\maketitle

\begin{abstract} Motivated by the need of the linking records across various databases, we propose a novel graphical model based classifier that uses a mixture of Poisson distributions with latent variables. The idea is to derive insight into each pair of hypothesis records that match by inferring its underlying latent rate of error using Bayesian Modeling techniques. The novel approach of using gamma priors for learning the latent variables along with supervised labels is unique and allows for active learning. The naive assumption is made deliberately as to the independence of the fields to propose a generalized theory for this class of problems and not to undermine the hierarchical dependencies that could be present in different scenarios. This classifier is able to work with sparse and  streaming data. The application to record linkage is able to meet several challenges of sparsity, data streams and varying nature of the data-sets. \end{abstract}

\section{Introduction} Data quality is one of the most important problems in data management, since dirty data often leads to inaccurate data analytic results and wrong business decisions. Poor data across businesses and the government cost the U.S. economy \$3.1 trillion a year, according to a report by InsightSquared in 2012 \cite{Ilyasxu}. In health care domains, keeping track of patients health information is vital and these data-sets reside in multiple data sources. All these records are critical to diagnose a disease or prescribe medicine for the disease and inaccurate or incorrect data may threaten patient safety \cite{kerrnorris}. Massive amounts of disparate data sources, have to be integrated and matched to support data analyses that can be highly beneficial to businesses, governments, and academia. 

Record Linkage is a process which aims to solve the task of merging records from different sources that refer to the same entity, a task that only gets harder if they don't share a unique identifier between them. The area of record linkage poses many challenges such as on-going linkage; storing and handling changing data; handling different linkage scenarios; accommodating ever increasing data-sets \cite{techchallenges}. All of these issues make the record linkage problem very challenging and critical. Efficient algorithms are essential to address this problem \cite{Mamun}.  
Traditionally, record linkage consists of two main steps: blocking and matching. In the blocking step, records that potentially match are grouped into the same block. Subsequently, in the matching step, records that have been blocked together are examined to identify those that match. Matching is implemented using either a distance function, which compares the respective field values of a record pair against specified distance thresholds, or a rule-based approach, e.g., “if the surnames and the zip codes match, then classify the record pair as matching" \cite{Karapiperis}.

The standard ways of solving such problems has been to use probabilistic model, alternative statistical models, searching and blocking, comparison and decision models \cite{Gu03recordlinkage}. One such method is the classical and popular Fellegi-Sunter method that estimates normalized frequency values and uses weighted combinations of them \cite{FellegiSunter}. This poses a problem as the underlying string error rates vary in a non-linear fashion and therefore these weighting schemes may not efficiently capture them. In addition, these linear weighting techniques are mostly ad-hoc and not based on a strong pattern recognition theory. Hence, automated linkage processes are one way of ensuring consistency of results and scalability of service. We propose a robust solution that models the probability of the matching records as a Poisson distribution that is learned using a probabilistic graphical model based classifier. 

Individual features have a latent error rate that are unique to them. For example, the errors in name pairs and errors in address pairs will likely have different error rates. We can learn such error rates by modeling them as gamma distributions which are conjugate priors to the Poisson likelihoods. The negative binomial distribution arises when the mixing distribution of the Poisson rate is a gamma distribution. The negative binomial distribution could be useful for various areas beyond record linkage where distribution of the features are Poisson, one example of such an application is RNA sequencing \cite{dong2016nblda}. This model allows for error to vary over time and produces better parameter estimates with increase in training data size which is especially useful in cases where you have streaming data or sparse data available for certain features.

\section{Probabilistic Record Linkage} The Fellegi-Sunter method for probabilistic record linkage calculates linkage weights which are estimated by observing all the agreements and disagreements of the data values of the variables that match. This weight corresponds to the probability that the two records refer to the same entity. Given two records ($R_a$, $R_b$), with a set of n common matching variables given by \begin{equation} R_a \rightarrow [F(a_1), F(a_2), ... ,F(a_n)] \end{equation} \begin{equation} R_b \rightarrow [F(b_1), F(b_2), ... ,F(b_n)] \end{equation}  the comparisons between the two records can be obtained as a result of applying a distance function like edit distance to each set of the matching variables and can be accumulated in a comparison vector 
\begin{equation} \alpha^c=\{ \alpha^c_1,\alpha^c_2, \ldots ,\alpha^c_n\} \end{equation} The binary comparison vector is calculated as 
\[
   \alpha^c_k=\left\{\begin{array}{lr}
       1, & \text{if}\ F(a_k) =F(b_k) \\
        0, & \text{otherwise }
       \end{array}\right\}
 \]

The basic idea in the Fellegi-Sunter method is to model the comparison vector as arising from two distributions – one that corresponds to true matching pairs and the other that corresponds to true non-matching pairs. For any observed comparison vector $\alpha^c$ in $\Lambda$ which is a space of all comparisons, the conditional probability of observing $\alpha^c$ given the pair is a match is given by $m(\alpha^c)=P(\alpha^c|(a,b)\ \epsilon \ {M}) $ and the conditional probability of observing $alpha$ given the pair is a non-match is given as $u(\alpha^c)=P(\alpha^c|(a,b) \ \epsilon \ U)$. Here $M$ and $U$ are the set of matches and the set of non-matches respectively. The weight for each record pair is given as $p_{ab} = \frac{m(\alpha^c)}{u(\alpha^c)}$ \cite{McVmurray}  \cite{sharp}. 

Once the probabilities are estimated, the decision rule for the Fellegi-sunter method is: if the weight of a record pair $p_{ab}$ \ $>$  \ $T_\lambda$, then its a match and if $p_{ab}$ \ $<$  \ $T_\tau$, then its a non-match. 

\subsection{Motivation} The standard mathematical technique of probabilistic record linkage in the state of the art, still relies on likelihood ratios and weights that are ad-hoc and cannot be probability measures. In addition the Fellegi-Sunter method calculates conditional probability by assuming a model which is very sensitive to original distribution in the overall database. In 2014, Toan Ong showed some good results in extending existing record linkage methods to handle missing field values in an efficient and effective manner \cite{Ong}. The supervised learning paradigms have been used by labeled comparison vectors but these suffer from the need to regularly updating training data \cite{Hurwitz}. A Bayesian approach to graphical record linkage was proposed which overcomes many obstacles encountered by previous approaches. This unsupervised algorithm is great for unlabeled data.  However, leveraging the vast amounts of labels when they are available is necessary. This begs the need for a truly probabilistic linkage theory and algorithm. Our algorithm is a truly probabilistic Poisson-gamma model that is popularly used in Bayesian statistics. We have applied this Poisson-gamma model to learn the latent error rates and hence proposed a Bayesian scheme for record linkage.  

In the big data era, the velocity of data updates is often high, quickly making previous linkage results obsolete. A true learning framework that can incrementally and efficiently update linkage results when data up-dates arrive are essential \cite{Gruenheid}. The advantage of this method is that it allows for only updating parameters in a active learning paradigm instead of training over all the large data-sets repeatedly as in majority of the state of the art, standard probabilistic record linkage algorithms. This would mean that a older data-sets can be thrown away and only the new data can be trained on and parameters updated. In a big-data setup this would be immensely useful

\subsection{Error Distribution} Given a pair of data-sets, the edit distances of pairs of strings within that data-sets that match, have relatively small edit distances as compared to the lengths of the matching string pairs. On the other hand, the edit distances of pairs of records are quite large compared to the lengths of the records in case of a non match. The total string length compared to the error rate are large and hence the error rates are modeled to be distributed as a Poisson. These errors that are distributed as Poisson can suffer from the uncertainty around the underlying error rate $\theta_i$ for the field $F_i$ which can be considered as a latent variable. 

The error distribution of the matches and non-matches are shown in Fig. 1:
\begin{figure}[htbp]
\centerline{\includegraphics[scale=0.55]{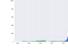}}
\caption{Error distribution of the Matches and Non-matches.}
\label{error_classes}
\end{figure}

Figures 2 and 3 shows the distribution of errors over the name and the address column which shows that the error rates could be different for different identifiers (features).

\begin{figure}[htbp]
\centerline{\includegraphics[scale=0.55]{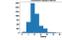}}
\caption{Error distribution of the name variable.}
\label{error_name}
\end{figure}

\begin{figure}[htbp]
\centerline{\includegraphics[scale=0.55]{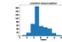}}
\caption{Error distribution of the address variable.}
\label{error_address}
\end{figure}

\subsection{Algorithm}

\begin{itemize} \item Using training data, infer parameters of the latent error rate that is modeled as a gamma distribution for each feature. \item Infer posterior predictive distribution over the test samples as a negative binomial distribution for the respective features. \item Compute the joint probability of the features and use training data to determine the optimum threshold for classification. \item Validate on test data. \end{itemize}
This algorithm is called the Negative Binomial Classifier and is further explained in the next section. 

\section{Supervised Negative Binomial Classifier} The errors though distributed as Poisson can suffer from the uncertainty around the underlying error rate $\theta_i$ for the field $F_i$.Hence we can consider the rate to be a probability distribution.We choose conjugate prior of Gamma distribution as a prior distribution for the latent variable $\theta_i$. Conjugate priors are chosen for mathematical convenience \cite{gelman2003bayesian}.

The naive independence assumption is for generality of the theory and by no means is the only formulation that it is limited by. The conjugation property enables learning of the underlying error rate and allows updating the latent variables as new data becomes available. The inference of the latent variable learning would constitute the learning and prediction would be the predictive distribution of the posterior probability considering the gamma prior and the Poisson likelihood.

\subsection{Poisson Distribution} The Poisson distribution is a convenient distribution to model the errors X with rate theta. The probability of an individual pair of records ($a_i$, $b_j$) which is a single observation is given by

\begin{equation} P(X|\theta)=\frac{\theta^x e^{-\theta}}{x!} \end{equation}

\subsection{Latent rate of error $\theta$:} The latent variable $\theta$ is the conjugate prior distribution which is known to be the Gamma distribution for the Poisson likelihood.\\

The error rate for a field or finite record pair is distributed as 

\begin{equation} p(\theta | y) = \gamma(\alpha,\beta) \end{equation}

The gamma distribution is a natural fit as a conjugate prior to the Poisson distribution \cite{gelman2003bayesian}.

$\alpha$ \& $\beta$ are parameters estimated from the ground truth. There are many techniques such as MLE, method of moment, etc to determine the parameters $\alpha$, $\beta$ \cite{minka2002estimating}.

\subsection{Estimation of $\alpha$, $\beta$:} We use method of moments to determine the parameter of the $\gamma$ distribution for a given feature $F_a$.

\begin{equation} E(X_i) = (\frac{\alpha}{\beta}) Var(X_i) = (\frac{\alpha}{\beta} + \frac{\alpha}{\beta^2}) \end{equation} Substituting E, \begin{equation} E(X_i) = \frac{\alpha}{\beta} and \end{equation} \begin{equation} Var(X_i) = E(.) (1+ \frac{1}{\beta}) = \frac{\alpha}{\beta} (1+\frac{1}{\beta}) \end{equation}

\subsection{Application to Record Linkage} \label{appl_rec_link} The conjugate prior along with the likelihood allows us to derive the posterior predictive distribution for the single pair of features $x_i (F(a_i),F(b_i))$ corresponding to the record pair ($R_a$, $R_b$) , given by: \\

\begin{equation} P((F_{a_i}, F_{b_i}) = \frac{P((F_{a_i}, F_{b_i}) | \theta_i))P(\theta_i)} {P(\theta | (F_{a_i}, F_{b_i})} \end{equation}

Writing this in terms of the variable $x_i$ we get, 

\begin{equation} P(x_i) = \frac{Poisson(X_i|\theta) Gamma(\theta_i|\alpha,\beta)}{Gamma(\theta_i|\alpha + X_i, 1+\beta)} \end{equation}

The parameter $\theta_i$, which is the rate of the Poisson distribution is distributed as Gamma:

\begin{equation} p(\theta_i) \sim \gamma(\alpha,\beta) \end{equation}

\begin{equation} P(x_i) = \frac{\gamma(\alpha + x_i) \beta^\alpha}{\gamma(\alpha) x_i! (1 + \beta)^{\alpha + x_i}} \end{equation}

This has a known form called the negative binomial distribution which is the posterior predictive distribution given the previously determined parameters. 

\begin{equation} P(x_i) = (\frac{\alpha + x_i -1}{x_i}) {(\frac{\beta}{\beta + 1}})^\alpha {(\frac{1}{\beta + 1}})^{x_i} \end{equation}
 \begin{equation} x_i \sim Neg-Bin (\alpha,\beta) \end{equation}

\subsection{Fuzzy Matching} To compare the probability of the records pairs ($R_a$, $R_b$), $R_a$, $R_b$ could potentially contain features that are not identical such as zip code present in one and not present in the other. We are however only looking at only the n fields that are common between the records. We shall now apply the generalized theory. This involves learning different error rates for each data-set. The assumption is that the error rate can vary for an individual pair of records with an underlying overall error rate.  Note that the latent parameters of the error distribution of only computed from the matching records and applied to the non-matching records.

The probability of the pair would be the joint of each feature. \begin{equation} P[ (F_{a_1},F_{b_1}), (F_{a_2},F_{b_2}), (F_{a_3},F_{b_3}), ..., (F_{a_n},F_{b_n}) ] \end{equation}
 Assuming the features are independent,
\begin{equation} 
\begin{split}
& = P(F_{a_1},F_{b_1}) \times P(F_{a_2},F_{b_2})\times ... \times P(F_{a_n},F_{b_n}) \\
& =  \prod_{i=1}^{n} (F_{a_i},F_{b_i})  \\
& =  \prod_{i=1}^{n} (\frac{\alpha_i + x_i -1}{x_i}) {(\frac{\beta_i}{\beta_i + 1}})^{\alpha_i} {(\frac{1}{\beta_i + 1}})^{x_i} 
\end{split}
\end{equation}
$\alpha_i, \beta_i \rightarrow$ 
parameters of the Gamma distribution $x_i \rightarrow$ variable $x_i$ for which the predictive distribution is applicable \\
\begin{equation} 
\begin{split}
& = (\frac{\alpha_1 + x_1 -1}{x_1})(\frac{\alpha_2 + x_2 -1}{x_2}) ... (\frac{\alpha_n + x_n -1}{x_n}) \\
& \times {(\frac{\beta_1}{\beta_1 + 1}})^{\alpha_1} {(\frac{\beta_2}{\beta_2 + 1}})^{\alpha_2} ... {(\frac{\beta_n}{\beta_n + 1}})^{\alpha_n} \\
& \times {(\frac{1}{\beta_1 + 1}})^{x_1} {(\frac{1}{\beta_2 + 1}})^{x_2} ... {(\frac{1}{\beta_n + 1}})^{x_n} \\
& = \prod_{i=1}^{n} Neg.Bin(\alpha_i,\beta_i,x_{i,j}) 
\end{split}
\end{equation}
 Assuming the pairs of data belongs to a class $c_k$, given k classes 
\begin{equation} H = \prod_{i=1}^{n}  Neg.Bin(\alpha_i,\beta_i,x_{i,j},c_k) \end{equation}
 Therefore the best class that the dataset fits is given by 
\begin{equation} H^k = \prod_{i=1}^{n} Neg.Bin(\alpha_i,\beta_i,x_{i,j},c_k) > {\theta} \end{equation}
 \qquad \qquad $k \ \epsilon(1,k)$ $\ x_{i,j}$ and threshold $\theta$\\
 The matching records can be chosen from the probability that crosses the threshold $\theta$. This above formulation should be useful in any scenario where features individually fit a Poisson distribution. The advantage of the Bayesian modeling approach is the robustness that the Negative Binomial distribution offers, due to the inherent conjugacy of the latent error rate. Not only is this approach generative in nature but also can learn from the ground truth. This can therefore be a combination of advantage of supervised learning and generative modeling.

\subsection{Active Learning} The parameters at each new data-set can be updated based on the fitness test as given by \begin{equation} \alpha_i = \alpha_i + Y_i \end{equation} \begin{equation} \beta_i = \beta_i + X_i \end{equation} Where $Y_i$ was the event for that occurred given $X_i$\\ Hence, $\theta_i|$($Y_i,X_i$) $\sim$ Gamma($\alpha_i + Y_i$, $\beta_i + X_i$) This can be represented as a graphical model shown in Fig 1. This classifier provides better estimates based on every past data that has been learned. Also, it may not be necessary to learn parameters at every step but in chunks depending on the application. In such an instance, a chunk of new data could be added to the stream and the underlying $\alpha$ and $\beta$ parameters can be recomputed. This will provide better estimates for the arriving test data set in the data stream.  

\section{Results}
This algorithm was applied to record linkage problem and was applied on restaurant data-sets. The restaurant data-sets are tables of names and addresses provided by the restaurant review companies  Zagat and Fodors. This is a real data containing real string data and real errors. This was downloaded from the RIDDLE data repository. There are a total of 191 records in the data-set out of which 60\% are unique.  The ground truth consists of false pairs of data sets along with matching pairs. This is a small data-set with only 113 true matches and 71 false matches. The performance of record linkages well as general classification algorithms suffer greatly with insufficient data. This data is divided in to training and testing data with 70\% used for training and rest is used to evaluate the performance of the algorithm. 

\subsection{Choosing the Threshold}
The parameters of the negative binomial distribution for each pair of features is learned from the matching records. For aggressive error biasing, the distribution of errors for the non matching pairs could be fit as a Gaussian normal. The matches were randomized with no criterion to filter the feature set on. This helps absorb random errors that could happen during the filtering operation as errors can happen at any string position. 
The log-likelihood of a record pair being a match is then calculated using the trained negative binomial distribution on all the records in the training data. The confusion matrix as well as the ROC and the Precision-Recall curves are then produced using the available labels on the training data. 

The ROC curve and the  Precision-Recall curve shown in figures 4 and 5 are then used to estimate the optimum threshold that can be used to make a decision about whether a pair of records are a match or not, on new data. 
One of the main advantages of this method which cannot be articulated enough is its ability to update parameters as and when new data is available. Let's say a new set pair of records needs to be matched, the likelihood of this pair is found using the negative binomial classifier and based on the threshold chosen, is deemed as either a match or a mismatch. If this turns out to be match, it can then further be used to update the parameters of the gamma distribution. The new data further solidifies the ability of the system. One of the other core advantages of this model is that the system does not need to be retrained over and over again and is averse to over-fitting due to its Bayesian nature. 

\begin{figure}[htbp]
\centerline{\includegraphics[scale=0.55]{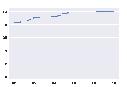}}
\caption{The AUC on the training data.}
\label{roc}
\end{figure}

\begin{figure}[htbp]
\centerline{\includegraphics[scale=0.55]{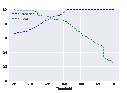}}
\caption{Precision - Recall curve on the training data.}
\label{precision-recall}
\end{figure}
As you can see from the figures 4 and 5, the model fits the training data very well evidenced by the high values of the AUC as well as Precision and Recall. But the utility of a model can only be judged by its performance on an unseen test data. The performance on the test data is as follows

\begin{tabular}{ l |  r }
  \hline			
  AUC & 94.23\%  \\
\hline
  Precision & 90.9\%  \\
\hline
  Recall & 76.92\% \\
 \hline  
  Accuracy & 86.2\% \\
 \hline
\end{tabular}
\label{table:Performance on the test data}

Even with a very small number of matching records that have been used to train the model, it is performing quite well on the test data. The very high AUC score shows that the negative binomial model is performing extremely well as a classifier. The precision of about 91\% also shows that the model is able to get a high percentage of results that are either match or non-match and also that a smaller number of actual matches are classified as non - matches (low false negative). This gives the confidence that if this model says that a pair of records do not match, then there is a very possibility that they actually don't. The overall accuracy of the model shows that 86\% of the records are classified properly as matches or non-matches. The threshold has been so chosen so as to focus more on precision than recall such that we would like to reduce the number of times the model identifies a match as a non-match.  In order to do this we would risk making some non-matches as matches.  

But due to the active learning framework of the model, the new arriving data are not only classified as match or non-match but are also used to improve the model. Thus with a larger data-set, we can achieve higher values of Precision, Recall as well as accuracy overall, even though we would still have to make a trade off between them.

\section{Conclusion and Future Work} This paper presents a classifier that is based on Bayesian principles and is robust and can work with small data-sets. This theory allows for active learning where additional data can be used to update the latent gamma distribution parameters. This theory provides a framework for researchers to explore other distributions as long as the underlying latent variables are conjugate priors. The excellent AUC score and accuracy is encouraging given the small size of the data. The Negative binomial classifier can be applied to different area of application as well like  RNA sequencing and other decay processes. The model can also be made hierarchical by assuming that the parameters of the gamma distribution ($\alpha$,$\beta$) follow some other probability distribution which would make them hyper-priors to the original error Poisson distribution.Researchers in record linkage can explore the variants of the theory to include hierarchical arrangements of features where dependencies such as zip code and street names can be set to further refine the model. Also, the target variable can be chosen as a mixing distribution of Dirichlet to account for imbalanced data-set.


\begin{thebibliography}{00}
\bibitem{techchallenges} J. H. Boyd, S. M. Randall, A. M. Ferrante, J. K. Bauer, A. P. Brown and J. B. Semmens,``Technical challenges of providing record linkage services for research,'' BMC Medical Informatics and Decision Making 14(1):23,  March 2014.
\bibitem{Karapiperis}  D. Karapiperis, A. Gkoulalas-Divanis and V. S. Verykios,  ``Summarization Algorithms for Record Linkage,"  EDBT, 2018. 
\bibitem{Mamun} A. A. Mamun, R. Aseltine and  S. Rajasekharan, ``Efficient Record Linkage Algorithms Using Complete Linkage Clustering," PLoS ONE 11(4): e0154446, 2016. 
\bibitem{Ilyasxu} I. F. Ilyas and X. Chu,``Trends in Cleaning Relational Data: Consistency and Deduplication,'' Foundations and Trends in Databases: Vol. 5: No. 4, pp. 281-393, 2015.
\bibitem{kerrnorris} K. Kerr, T. Norris, and R. Stockdalel, ``Data quality information and decision making: a healthcare case study", In Proceedings of the 18th Australasian Conference on Information Systems Doctoral Consortium, pages 5–7, 2007. 
\bibitem{FellegiSunter} I. P. Fellegi and A. B. Sunter, ``A theory for record linkage,'' Journal of the American Statistical Association, 64(328):1183--1210, 1969.
\bibitem{dong2016nblda} K.Dong, H. Zhao, T. Tong and X. Wan, ``Nblda: negative binomial linear discriminant analysis for rna-seq data,'' BMC bioinformatics, 17(1):369, 2016. 
\bibitem{Gu03recordlinkage} L. Gu, R. Baxter, D. Vickers and C. Rainsford, `` Record linkage: Current practice and future directions," Technical report, CSIRO Mathematical and Information Sciences, 2003. 
\bibitem{gelman2003bayesian} A. Gelman, J. Carlin, H. Stern and D. Rubin, Bayesian Data Analysis, Second Edition. Chapman \&  Hal  Texts in Statistical Science, 2003. 
\bibitem{McVmurray} B. S. McVeigh and J. S. Murray, ``Practical Bayesian Inference for Record Linkage," Technical report, Carnegie Mellon University, 2017. 
\bibitem{sharp}  S. Sharp, ``Deterministic and probabilistic record Linkage," Alternative sources branch, National Records of Scotland. 
\bibitem{minka2002estimating}  T. P. Minka, ``Estimating a gamma distribution,"Microsoft Research, Cambridge, UK, Tech. Rep, 2002. 
\bibitem{Gruenheid}  A. Gruenheid, X. L. Dong and  D. Srivastava, ``Incremental Record Linkage,"Proc. VLDB Endow, Vol 7:No. 9, pp  697--708, May 2014. 
\bibitem{Ong}  T. C. Ong, M. V. Mannino, L. M. Schilling and M. G. Kahn, ``Improving record linkage performance in the presence of missing linkage data," Journal of Biomedical Informatics, Vol 52, pp 43-54, December 2014. 
\bibitem{Hurwitz}  A. M, Hurwitz,  `` Record linkage sharing using labeled comparison vectors and a machine learning domain classification trainer,"  US Patent, US9576248B2.  
\end{thebibliography}
\end{document}